# A multimodal vision foundation model for generalizable knee pathology

Kang Yu[1,2#], Dingyu Wang[3,4,5#], Zimu Yuan[3,4,5#], Nan Zhou[1,2#], Jiajun Liu[3,4,5], Jiaxin Liu[3,4,5], Shanggui Liu[3,4,5], Yaoyan Zheng[1,2], Huishu Yuan[6*], Di Huang[1,2*], Dong Jiang[3,4,5*]

1. State Key Laboratory of Complex and Critical Software Environment, Beihang University, Beijing, China
2. School of Computer Science and Engineering, Beihang University, Beijing, China
3. Department of Sports Medicine, Peking University Third Hospital, Institute of Sports Medicine of Peking University, Beijing, China.
4. Beijing Key Laboratory of Research and Translation for Drugs and Medical Devices in Precision Diagnosis and Treatment of Sports Injuries, Beijing, China.
5. Engineering Research Center of Sports Trauma Treatment Technology and Devices，Ministry of Education, Beijing, China.
6. Department of Radiology, Peking University Third Hospital, 49 Huayuan North Road, Haidian District, Beijing, China.

\# These authors contributed equally to this study.

*These are corresponding authors.

E-mail: huishuy@bjmu.edu.cn; dhuang@buaa.edu.cn; bysyjiangdong@126.com

## Abstract

Musculoskeletal disorders, particularly those affecting the knee, represent a leading cause of global disability, creating an urgent demand for precise interpretation of medical imaging. Current artificial intelligence (AI) approaches in orthopedics predominantly rely on task-specific, supervised learning paradigms. These methods are inherently fragmented, require extensive annotated datasets, and often lack generalizability across different modalities and clinical scenarios. The development of foundation models in this field has been constrained by the scarcity of large-scale, curated, and open-source musculoskeletal datasets.

To address these challenges, we introduce OrthoFoundation, a multimodal vision foundation model optimized for generalizable musculoskeletal pathology analysis. We constructed a massive pre-training dataset comprising approximately 1.2 million unlabeled knee X-ray and MRI images, aggregated from the internal PUTH-Group dataset and public databases (OAI and FastMRI). Utilizing a Dinov3-Large backbone identified through rigorous transferability estimation, the model was trained via self-supervised contrastive learning to capture robust radiological representations. We evaluated the model's performance using full fine-tuning and linear probing across a comprehensive benchmark of downstream tasks.

OrthoFoundation achieves or exceeds state-of-the-art (SOTA) performance across 14 distinct downstream tasks involving diagnosis, disease progression, and prognosis. In X-ray-based osteoarthritis diagnosis, it attained superior accuracy, while in MRI-based structural injury detection, it consistently ranked first. For instance, in the complex diagnosis of posterior cruciate ligament (PCL) tear, its AUC value reached 84.3%. The model demonstrated remarkable label efficiency, matching or surpassing fully supervised baselines using only 50% of the labeled training data. Furthermore, despite being pre-trained exclusively on knee images, OrthoFoundation exhibited exceptional cross-anatomy generalization, successfully identifying pathologies in the hip, shoulder, and ankle.

OrthoFoundation represents a significant advancement toward general-purpose AI for musculoskeletal imaging. By learning fundamental, joint-agnostic radiological semantics from large-scale multimodal data, it overcomes the limitations of conventional supervised models. This work provides a robust framework for reducing annotation burdens and enhancing diagnostic precision in clinical orthopedic practice.

## Introduction

Musculoskeletal disorders represent a pervasive and debilitating global health challenge, affecting billions and constituting a leading cause of chronic pain, functional impairment, and disability worldwide[1,2]. Within this spectrum, the knee joint stands out as a site of exceptional clinical concern. As the largest and one of the most complex joints, it is uniquely vulnerable, bearing the brunt of acute sports-related injuries, from meniscal tears to ligamentous ruptures, and the chronic, degenerative ravages of osteoarthritis, which is itself a primary driver of global musculoskeletal disability[3,4]. The high prevalence and severe impact of these knee conditions create an urgent and continuous demand for precise diagnostic interpretation. Consequently, medical imaging, particularly through the widely utilized modalities of MRI for detailed soft-tissue assessment and X-ray for bony structural evaluation, is absolutely fundamental to guiding effective diagnostic and treatment pathways and improving patient outcomes[5].

In response to this need, AI has emerged as a powerful tool for medical image analysis[6]. In recent years, numerous deep learning models have been developed that demonstrate high proficiency in narrow, isolated tasks, such as detecting anterior cruciate ligament (ACL) ruptures or grading osteoarthritis severity[7,8]. However, this conventional paradigm is inherently fragmented: these models typically rely on task-specific architectures and supervised training from scratch using extensive annotated datasets. This approach results in rigid systems with poor generalizability, ultimately limiting their robustness and practical utility in the dynamic clinical environment[9].

A transformative shift in AI is underway with the advent of foundation models[10]. Trained at an unprecedented scale on broad, often unlabeled, datasets using self-supervised learning, these models learn fundamental representations and structures inherent in multimodal data. Their key strength is an emergent ability to perform a diverse array of downstream tasks through techniques like prompting or minimal fine-tuning, moving away from the one-model-one-task paradigm[11]. This architecture promises to significantly enhance the efficiency, robustness, and generalizability of AI systems, as evidenced by their successful deployment in fields like ophthalmology[12], pathology[13], and dermatology[14]. Despite this transformative potential, the development of foundation models in orthopedics has notably lagged behind, primarily due to a critical lack of large-scale, curated, and open-source datasets necessary for effective pre-training. Additionally, the absence of a unified evaluation benchmark covering MSK data limits the effective validation and benchmarking of model performance.

To address these challenges, we present OrthoFoundation, a foundation model for diagnosis and prognosis of knee diseases on X-ray and MRI. We first constructed PUTH-Group dataset with 414406 knee X-rays and 974893 knee MRI images from three centers of PUTH for pretraining, fine-tuning and validation. Together with public osteoarthritis initiative (OAI) dataset[15] and fastMRI[16], we trained OrthoFoundation on a dataset containing 1.2 million X-ray and MRI images. With minimal labeled data for fine-tuning, Orth-Foundation achieves or exceeds state-of-the-art (SOTA) performance across 14 downstream tasks involved with diagnosis and prognosis. Furthermore, the model exhibits exceptional generalization ability to hip, ankle and should diseases.

## Results

### Overview

Orthofoundation is a multimodal vision pretrained model, trained with 1.2 million images of knee X-ray and MRI. The data details can be found in Figure 1a and Table 1. By the Orthofoundation utilized Dino structure, specifically the DINOv3-Large backbone, it employs a self-supervised student-teacher framework. Through contrastive learning and multi-crop augmentation strategies, the model captures both gross skeletal structures from radiographs and fine-grained soft-tissue details in MRI scans, establishing a unified and robust feature representation space. Following training, Orthofoundation can be finetuned in a data-saving manner for downstream applications such as knee pathology diagnosis and KOA prediction.

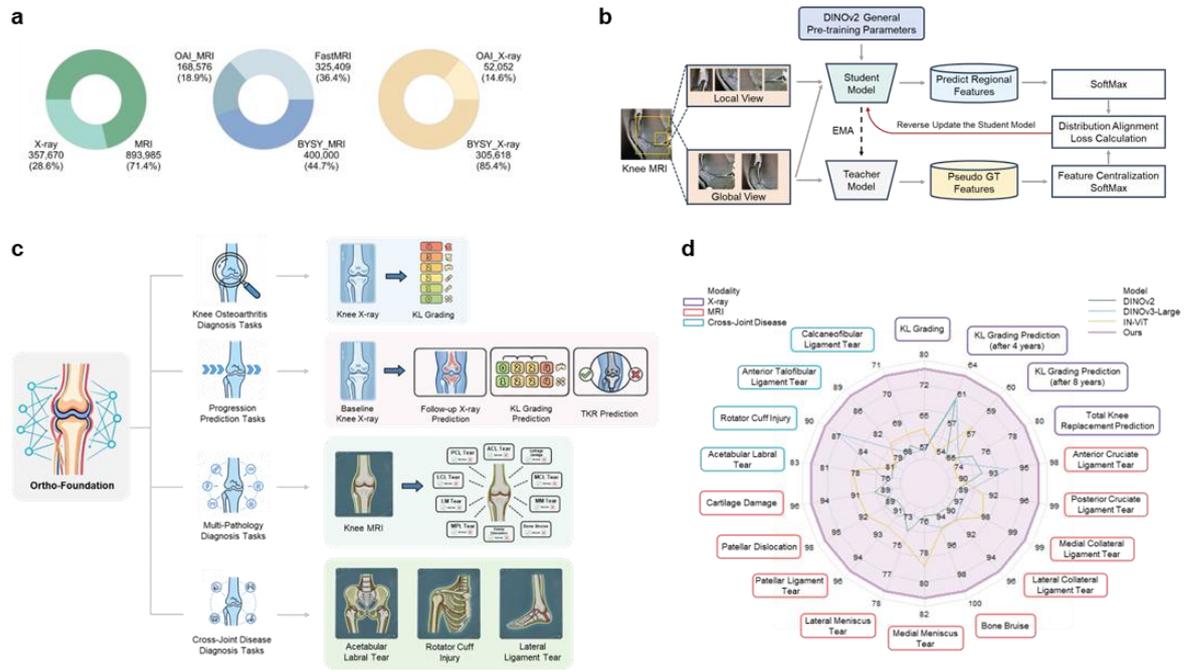

Figure 1: Overview of this study.

Table 1: Overview of the data used in our study.

| Modal | Usage | Downstream Tasks | | Volume |
|---|---|---|---|---|
| X-ray | Pre-training | - | | 357670 |
| | Fine-tuning | Knee Osteoarthritis Diagnosis Task | KL Grading | 46057 |
| | | Knee Osteoarthritis Progression Prediction Tasks | KL Grading Prediction (4 years follow-up) | 6331 |
| | | | KL Grading Prediction (8 years follow-up) | 3657 |
| | | | TKR prediction | 691 |
| | Total | | | 414406 |
| MRI | Pre-training | - | | 893985 |
| | Fine-tuning | Knee Multi-Pathology Diagnosis Tasks | ACL Tear | 2422 |
| | | | PCL Tear | 10760 |
| | | | MCL Injury | 11021 |
| | | | LCL Injury | 23596 |
| | | | Meniscus Tear | 5467 |
| | | | Patellar Dislocation | 6207 |
| | | | Medial Patellar Retinaculum Injury | 4898 |
| | | | Bone Bruise | 1271 |
| | | | Cartilage Damage | 5846 |
| | | Cross-Joint Injury Diagnosis Tasks | Acetabular Labral Tear | 4584 |
| | | | Rotator Cuff Injury | 2688 |

| | | |
|---|---|---|
| | Anterior Talofibular Ligament Tear | 1074 |
| | Calcaneofibular Ligament Tear | 1074 |
| Total | | 974893 |

**Model Selection & Pretraining Strategy**

To construct a foundational backbone optimized for musculoskeletal imaging, we first conducted a rigorous transferability estimation analysis across ten candidate architectures. Instead of arbitrarily selecting a model backbone, we systematically evaluated the adaptability of each architecture across five distinct medical imaging datasets (Supplementary Table 1), balancing performance potential with computational cost. Based on aggregated transferability scores and parameter efficiency, Dinov3-Large was identified as the optimal architecture.

Building on this, we developed OrthoFoundation, a multimodal foundation model pre-trained on a large-scale dataset. The model employs self-supervised contrastive learning and was trained on approximately 1 million unlabeled knee images (X-ray and MRI) from the OAI, FastMRI, and Peking University Third Hospital databases. As shown in Fig.1, OrthoFoundation demonstrated superior scalability and more efficient convergence compared to conventional pre-training paradigms. Through this training process, the resulting representations of OrthoFoundation capture both gross skeletal structures from radiographs and fine-grained soft-tissue details in MRI scans, thereby establishing a unified feature representation space for downstream clinical tasks.

**Comprehensive diagnostic performance across modalities**

We systematically evaluated the diagnostic performance of OrthoFoundation in both X-ray and MRI modalities using full fine-tuning. For X-ray-based KOA diagnosis, OrthoFoundation achieved SOTA performance on the 5-class KL grading task. On the internal test set, our model attained an accuracy of 77.8% and a macro-F1 score of 79.2.

We further extended the evaluation to MRI-based structural injury diagnosis, a task requiring precise interpretation of soft-tissue signals. Across ten binary classification tasks, including the anterior/posterior cruciate ligaments (ACL/PCL), medial/lateral menisci, and collateral ligaments, patellar dislocation, bone bruise, medial patellar retinaculum, and cartilage damage, OrthoFoundation consistently ranked first. Notably, for the challenging diagnosis of PCL tears, our model achieved an AUC of 84.3, surpassing the second-best model (In-ViT) by four percentage points. Heatmap visualizations confirmed that OrthoFoundation accurately focused on lesion regions, such as ligament discontinuity and high-signal intensity areas, closely aligning with the attention patterns of radiologists (Fig. 2c).

**Long-term prognosis and disease progression prediction**

In addition to disease diagnosis, a critical capability of medical foundation models lies in predicting disease progression. We evaluated the performance of OrthoFoundation in two prognostic tasks: predicting KOA progression at 4-year and 8-year follow-up (as measured by KL grade changes) and predicting the necessity for Total Knee Replacement (TKR). As illustrated in Fig. 3a, our model demonstrated strong predictive performance, achieving a Macro-F1 score of 63.4 for 4-year progression and 58.5 for 8-year progression using only baseline X-ray images.

Crucially, in the TKR prediction task, our model successfully stratified patients into those who required replacement and those who did not. Compared to DinoV3-Large, the accuracy of our model in predicting TKR can be improved by 4-7 percentage points. These findings suggest that OrthoFoundation captures subtle radiographic features indicative of future joint degeneration that may be imperceptible to human observers.

**Cross-Anatomy Generalization**

To assess whether OrthoFoundation learned universal musculoskeletal features rather than knee-specific patterns, we evaluated its generalization capabilities on hip, shoulder and ankle diseases. Despite not being exposed to non-knee images during pretraining, OrthoFoundation exhibited remarkable transferability. Specifically, in detecting hip labral tears, rotator cuff injuries, as well as anterior talofibular ligament (ATFL) and calcaneofibular ligament (CFL) injuries, our model achieved AUCs of 83.7, 86.1, 86.9 and 70.6 respectively (Fig. 4). When compared to three baseline models, Dinov2, Dinov3, and In-ViT, our model showed consistent superiority, achieving average performance gains of 3-4 percentage points. This robust cross-anatomy generalization indicates that our pretraining strategy effectively encoded fundamental radiological semantics of bone and soft tissue injuries, facilitating rapid adaptation to novel orthopedic tasks.

**Label Efficiency**

Medical AI development is often bottlenecked by the scarcity of expert annotations. We evaluated the label efficiency of all models by fine-tuning on subsets of training data (12.5%, 25%, 50%, 100%). As shown in Fig. 4, OrthoFoundation maintained robust performance even in data-limited regimes. Remarkably, with only 50% of the labeled data, our model achieved an AUC of 97.7 for ACL tear detection, surpassing the performance of Dinov3 trained on 100% of the data. This high label efficiency suggests that OrthoFoundation can significantly reduce the annotation burden for deployment in new clinical settings.

## Discussion

In this study, we developed Ortho-Foundation, a multimodal vision foundation model for musculoskeletal imaging. Our results demonstrate that the model is highly adaptable to a diverse spectrum of musculoskeletal diseases, extending from knee-specific pathologies such as osteoarthritis, ligament tears, and meniscal injuries, to conditions affecting other major joints including the hip, shoulder, and ankle. Through extensive evaluation across seventeen distinct downstream tasks involving diagnosis, prognosis, and cross-joint generalization, Ortho-Foundation achieved or surpassed state-of-the-art performance, underscoring its robust and generalizable representations learned from large-scale, multimodal knee imaging data.

A key bottleneck in the development of medical foundation models has been the scarcity of large, curated, and openly available imaging datasets required for effective self-supervised pre-training[11,17]. Prior efforts in other medical domains have often been constrained by data volume and diversity. For example, early self-supervised frameworks like Models Genesis or Transfusion were primarily validated on relatively small-scale public datasets or single modalities[18,19], limiting their scalability to diverse clinical pathologies. In this work, we addressed this fundamental challenge by constructing a substantial pretraining dataset comprising approximately 1.2 million knee X-ray and MRI images, aggregated from multiple public and internal sources. Furthermore, we curated and annotated a substantial collection of images for downstream task fine-tuning, enabling rigorous validation across a wide array of clinical applications. This large-scale, multimodal dataset represents a significant resource for the musculoskeletal imaging community and serves as a cornerstone for training a model with broad generalizability.

While Ortho-Foundation is not the first foundation model applied to biomedical data[14,20,21], it is, to our knowledge, the first such model specifically designed and validated for the musculoskeletal system. Compared to conventional supervised models that are typically designed for single, narrow tasks[7,22], the strength of our foundation model lies in its inherent capacity for multi-task learning and multimodal generalization[10]. Importantly, although the model was pre-trained exclusively on knee images, it exhibited remarkable transferability to diseases of the hip, shoulder, and ankle. This indicates that Ortho-Foundation learned fundamental, joint-agnostic radiological semantics of bone and soft-tissue pathology, which facilitates efficient adaptation to novel diagnostic tasks with minimal labeled data. This capability significantly reduces the annotation burden and accelerates deployment in new clinical scenarios[23].

Despite its promising performance, this work has several important limitations. A primary limitation is the use of a 2D image encoder architecture, which processes individual MRI slices independently. This approach is not inherently suited for capturing the full spatiotemporal context available in 3D MRI volumes or dynamic sequences, potentially overlooking valuable information across adjacent slices or time points[24]. Future iterations would benefit from incorporating 3D or video-based architectures to better model the volumetric nature of medical imaging data[25,26]. The downstream tasks are mainly binary classification, but the real clinical demands are usually multi-classification and multi-stage tasks. Prospective clinical validation in real-world settings is necessary to fully ascertain the model's utility and integration into routine diagnostic workflows[27].

In summary, Ortho-Foundation represents a significant step toward general-purpose AI for musculoskeletal imaging. By leveraging large-scale self-supervised learning on multimodal knee data, the model achieves high performance, label efficiency, and notable generalization across tasks and anatomy. We believe this work provides a foundational framework and a valuable benchmark for future research in orthopedic artificial intelligence, with the potential to improve diagnostic accuracy, prognostic insight, and ultimately patient care in musculoskeletal medicine.

## Methods

### Dataset

#### Dataset for Self-Supervised Pre-training

To enable the learning of robust and generalizable visual representations, we constructed a large-scale, unlabeled knee imaging dataset comprising 1,000,000 images for self-supervised pre-training. This dataset aggregates images from public databases and the internal datasets we have collected.

Public Datasets. We collected a total of 600,000 knee radiographic images from two international, publicly available databases.

Osteoarthritis Initiative (OAI). This is a multi-center, longitudinal, prospective observational study of knee osteoarthritis sponsored by the National Institutes of Health (NIH). From this resource, we obtained 100,000 knee X-ray images and 200,000 knee MRI slices, which served as a diverse source of imaging data from a well-defined patient cohort.

FastMRI. As a collaborative project between NYU Langone Health and Facebook AI Research, this dataset was developed to accelerate MRI reconstruction. We leveraged 300,000 knee MRI slices from this database, further enriching our pre-training data with images acquired using different protocols and hardware.

Internal Dataset. We retrospectively collected 400,000 knee images from three medical centers (Peking University Third Hospital-Main Center, Peking University Third Hospital-Beijing Capital International Airport Center, Peking University Third

Hospital-North Center) between December 2016 and June 2023. This internal dataset consists of 200,000 X-ray images and 200,000 MRI slices, reflecting our local patient demographics and clinical imaging equipment.

Data Curation and Preprocessing. To ensure data consistency and quality, we have adopted a standardized preprocessing procedure. All images were initially exported in DICOM format and subsequently converted to PNG format for compatibility with our model framework. To minimize noise, a quality control audit was conducted by a team of five orthopedic surgeons and musculoskeletal radiologists. Through random sampling, they reviewed the dataset and excluded non-knee images, severely blurry or over/under-exposed images, stitched images, and those annotated with additional graphics. This comprehensive pre-training dataset encompasses a wide spectrum of knee pathologies, including but not limited to osteoarthritis, ligamentous injuries, meniscal tears, and chondral defects, covering more than ten common knee diseases.

**Dataset for downstream Tasks**

To assess the performance of our pre-trained foundation model, we established distinct downstream datasets for both X-ray and MRI modalities, each designed to evaluate specific clinical tasks.

X-ray Modality. We constructed a longitudinal cohort from the OAI database, comprising 45,000 knee X-ray images. We selected patients who had radiographs available at three distinct time points: baseline, 48-month follow-up (4 years), and 96-month follow-up (8 years). For each image, the corresponding Kellgren-Lawrence (KL) grade was extracted from OAI's annotations. Additionally, we extracted the final clinical outcome for each patient, defined as whether they eventually underwent total knee arthroplasty (TKA). This dataset was utilized to evaluate the model on three clinical tasks: (1) classification of the KL grade from a given X-ray image; (2) prediction of KL grade progression at 48- and 96-month follow-ups based on the baseline X-ray; and (3) prediction of the long-term risk of requiring TKA, also based on the baseline image.

MRI Modality. We established a large-scale, multi-center MRI dataset comprising 200,000 images from 18,726 patients across three different medical centers. This dataset covers ten common knee pathologies: anterior cruciate ligament (ACL) tear, posterior cruciate ligament (PCL) injury, medial meniscus (MM) tear, lateral meniscus (LM) tear, medial collateral ligament (MCL) injury, lateral collateral ligament (LCL) injury, bone contusion, medial patellofemoral ligament (MPFL) injury, patellar dislocation, and cartilage injury. Patients who had undergone knee surgery prior to the MRI examination or whose images had significant artifacts were excluded. All images were exported from the Picture Archiving and Communication System (PACS) in DICOM format and converted to PNG.

The dataset includes proton density-weighted fat suppression (PDWFS) sequences in the sagittal, coronal and axial planes. The images were acquired on scanners from multiple manufacturers (e.g., Siemens, GE, Philips) with field strengths of 1.5T and 3.0T, reflecting real-world clinical heterogeneity.

We implemented a rigorous annotation process to ensure label accuracy, which was performed by three board-certified musculoskeletal radiologists, each with over 10 years of experience. To mitigate bias, annotators were blinded to the original radiological reports and independently performed binary classification (normal/abnormal) for each of the 10 anatomical structures or pathologies, with any diagnostic discrepancies resolved through joint discussion and majority vote. To establish the highest possible label fidelity, we have adopted different gold standards: for ACL tears and meniscal tears in patients who later underwent arthroscopy, intraoperative surgical findings served as the ground truth, while for all other eight pathologies, the consensus of the three senior radiologists was used as the ground truth. We used a five-fold cross-validation strategy to evaluate the model. The dataset was divided into five folds at the patient level. In each iteration, four folds were used for training and one for validation, reporting the average performance across all five folds.

**Dataset for Label-Efficient Evaluation**

To assess the model's data efficiency and its performance in low-data conditions, we created a subset from the downstream task MRI dataset. We specifically focused on four common conditions: ACL tear, PCL injury, MCL injury, and cartilage injury. For each condition, we fine-tuned the model using randomly sampled fractions of the complete training set (25%, 50%, 75%, and 100%). Then evaluate the performance on the test set to determine the model's learning capability using different amounts of labeled data.

**Dataset for Cross-Domain Generalization**

To evaluate the model's ability to generalize to distinct joints and thereby test its robustness and transfer learning capabilities, we collected three additional MRI datasets. The annotation process followed the same expert consensus protocol as the knee MRI dataset in downstream task evaluation.

We retrospectively collected MRIs from 605 patients with hip pathologies (acetabular labral tear), 2,688 patients with shoulder pathologies (supraspinatus tendon injury), and 1,753 patients with ankle pathologies (anterior talofibular and calcaneofibular ligament injuries). The downstream tasks included binary classification (normal/abnormal) for the hip labrum and ankle ligaments, and tertiary classification (normal/degeneration/tear) for the supraspinatus tendon. Each dataset was split into a training set (80%) and a test set (20%) at the patient level. The model was fine-tuned on the training set and evaluated on the test set to assess its cross-domain performance.

**Estimation of transferability**

To select an optimal pre-trained backbone for subsequent continued pre-training on our target knee imaging datasets, we conducted a comprehensive transferability estimation analysis. This step was designed to identify the model with the highest potential for adaptation to medical imaging tasks, balancing performance potential with computational cost.

We evaluated a panel of state-of-the-art pre-trained Vision Transformer (ViT) models, including DINO-B, DINOv2-B, DINOv2-L, DINOv2-G, DINOv3-B, DINOv3-L, MAE-B, MAE-L, MoCov3-B, and SimMIM-B.

The transferability of these candidate models was assessed using a suite of established estimation methods, including PED, SFDA, ITM, NLEEP, LogME, and

PARC. These methods provide a proxy for downstream performance without necessitating full-scale model fine-tuning.

We benchmarked the models on five diverse medical downstream datasets, spanning multiple imaging modalities and pathologies: (1) eye disease classification from fundus images, (2) pneumonia recognition from chest X-rays, (3) pneumothorax detection from chest X-rays, (4) leukemia classification from histopathology slides, and (5) melanoma detection from dermoscopic images.

Each model's performance potential on these tasks was scored using the aforementioned transferability estimation methods. By aggregating the scores across all tasks and estimation techniques, we derived a comprehensive ranking. After synthesizing the overall transferability scores and considering the trade-off with model size (i.e., parameter count), DINOv3-L was selected as the optimal base model for our subsequent continued pre-training phase.

**Model architecture and pretraining**

Our model utilizes a Vision Transformer (ViT) backbone, initialized with DINOv2 weights to leverage its powerful general visual representations. The core of our approach was to adapt foundation model for domain-specific expertise through continued self-supervised pretraining on a curated, million-scale dataset of knee images. We employed the DINO student-teacher framework, where a student network learns by matching the output of a teacher network. The teacher's weights are an exponential moving average (EMA) of the student's, providing stable pseudo-labels. Following the established multi-crop augmentation strategy, the student network processes both local and global views of an image, while the teacher observes only the global view. In doing so, the model is compelled to learn representations that are robust to changes in scale and viewpoint by minimizing the cross-entropy loss between the student's predictions and the teacher's centered and subsequently sharpened outputs, all without relying on manual annotations.

**Linear probing and full fine-tuning evaluation**

Following large-scale continual pre-training on knee imaging datasets, we further assessed the adaptability of the foundation model through two representative paradigms:

linear probing (LP) and full fine-tuning (FT).

Linear probing provides a low-cost and standardized means of evaluating the intrinsic representational quality of a pretrained model. In this setting, all model parameters are frozen, and only a linear classifier is trained on the extracted global embeddings. By restricting the evaluation to the feature space, LP directly measures the model's ability to capture clinically relevant patterns such as narrowing of the joint space, degenerative cartilage changes, cruciate and collateral ligament disruptions, or signs of patellar malalignment. Strong gains under LP compared with baseline models suggest that continued pre-training enhances the feature space in ways directly relevant to knee disease interpretation. Full fine-tuning, in contrast, updates all model parameters together with the task-specific classification head. This approach not only preserves the pretrained representations but also allows them to adapt to the distributional characteristics of new datasets, thereby maximizing task-specific diagnostic performance.

LP and FT serve complementary roles in evaluating foundation models. LP offers a rapid and resource-efficient assessment of feature transferability, while FT provides a more definitive test of adaptability to clinical endpoints. Together, these paradigms deliver a comprehensive framework for verifying both the generalizability and the task-specific refinement of the model, thereby informing its readiness for deployment in large-scale, multi-center clinical workflows.

**Evaluation metrics**

To comprehensively evaluate the performance of the proposed models, we employ specific metrics based on the type of classification task. For binary classification tasks, we utilized the area under the receiver operating characteristic curve (AUROC) and the area under the precision-recall curve (AUPR). For multi-class classification tasks, we used Accuracy, Macro-averaged F1-score (Macro-F1), and Weighted-averaged F1-score (Weighted-F1) for evaluation.

**Statistical analysis**

All experiments were evaluated using 5-fold cross-validation. Performance metrics are reported as mean ± standard deviation (SD). To assess the statistical significance of the performance improvements, we conducted two-sided t-tests between the proposed model and the baseline methods. P < 0.05 was regarded as statistically significant. All statistical analyses were performed by Python.

**Data availability**

The raw data collected and processed in this study are available under restricted access, which can be obtained by emailing the corresponding author with all requests for academic use.

**Code availability**

The pipeline development and experiments are conducted in Python with PyTorch as a primary tool. All code for reproducing this study can be found at https://github.com/ytrsk/OrthoFoundation.